\documentclass[10pt,twocolumn,letterpaper]{article}

\usepackage{iccv}
\usepackage{times}
\usepackage{epsfig}
\usepackage{graphicx}
\usepackage{amsmath}
\usepackage{amssymb}
\usepackage{booktabs}
\usepackage{subcaption}

\usepackage[pagebackref=true,breaklinks=true,letterpaper=true,colorlinks,bookmarks=false]{hyperref}

\iccvfinalcopy 

\def\iccvPaperID{2603} 

\ificcvfinal\fi
\begin{document}

\title{AVT: Unsupervised Learning of Transformation Equivariant Representations by Autoencoding Variational Transformations}

\author{Guo-Jun Qi$^{1,2,}$\thanks{Corresponding author: G.-J. Qi. Email: guojunq@gmail.com. The idea was conceived and formulated by G.-J. Qi, and L. Zhang performed experiments while interning at Huawei Cloud.}~, Liheng Zhang$^1$, Chang Wen Chen$^3$, Qi Tian$^4$\vspace{1.5mm}\\
$^1$Laboratory for MAchine Perception and LEarning (MAPLE)\\
\url{http://maple-lab.net/}\vspace{1mm}\\
$^2$Huawei Cloud, $^4$Huawei Noah's Ark Lab\\
$^3$ The Chinese University of Hong Kong and University at Buffalo\\
{\tt\small guojun.qi@huawei.com}\\
{\tt \small\url{http://maple-lab.net/projects/AVT.htm}}
}

\maketitle

\begin{abstract}
The learning of Transformation-Equivariant Representations (TERs), which is introduced by Hinton et al. \cite{hinton2011transforming}, has been considered as a principle to reveal visual structures under various transformations. It contains the celebrated Convolutional Neural Networks (CNNs) as a special case that only equivary to the translations. In contrast, we seek to train TERs for a generic class of transformations and train them in an {\em unsupervised} fashion. To this end, we present a novel principled method by Autoencoding Variational Transformations (AVT), compared with the conventional approach to autoencoding data. Formally, given transformed images, the AVT seeks to train the networks by maximizing the mutual information between the transformations and representations. This ensures the resultant TERs of individual images contain the {\em intrinsic} information about their visual structures that would equivary {\em extricably} under various transformations in a generalized {\em nonlinear} case. Technically, we show that the resultant optimization problem can be efficiently solved by maximizing a variational lower-bound of the mutual information. This variational approach introduces a transformation decoder to approximate the intractable posterior of transformations, resulting in an autoencoding architecture with a pair of the representation encoder and the transformation decoder. Experiments demonstrate the proposed AVT model sets a new record for the performances on unsupervised tasks, greatly closing the performance gap to the supervised models.
\end{abstract}

\section{Introduction}\label{sec:intro}
Convolutional Neural Networks (CNNs) have demonstrated tremendous successes when a large volume of labeled data are available to train the models.
Although a solid theory is still lacking, it is thought that both {\em equivalence} and {\em invariance} to image translations play a critical role in the success of CNNs \cite{cohen2016group,cohen2018intertwiners,sabour2017dynamic,hinton2011transforming}, particularly for supervised tasks.

Specifically, while the whole network is trained in an end-to-end fashion,
a typical CNN model consists of two parts: the convolutional {\em feature maps} of an input image through multiple convolutional layers, and the classifier of {\em fully connected layers} mapping the feature maps to the target labels.
It is obvious that a supervised classification task requires the fully connected classifier to predict labels invariant to transformations. For training a CNN model, such a transformation invariance criterion is achieved by minimizing the classification errors on the labeled examples
augmented with various transformations \cite{krizhevsky2012imagenet}. Unfortunately, it is impossible to simply apply transformation invariance to learn an unsupervised representation without label supervision, since this would result in a trivial constant representation for any input images.

On the contrary, it is not hard to see that the representations generated through convolutional layers are equivariant to the transformations -- the feature maps of translated images are also shifted in the same way subject to edge padding effect \cite{krizhevsky2012imagenet}. It it natural to generalize this idea by considering more types of transformations beyond translations, e.g., image warping and projective transformations \cite{cohen2016group}.

In this paper, we formalize the concept of {\em transformation equivariance} as the criterion to train an unsupervised representation. We expect it could learn the representations that are generalizable to unseen tasks without knowing their labels. This is contrary to the criterion of transformation invariance in supervised tasks, which aims to tailor the representations to predefined tasks and their labels.
Intuitively, training a transformation equivariant representation is not surprising -- a good representation should be able to preserve the {\em intrinsic} visual structures of images, so that it could {\em extrinsically} equivary to various transformations as they change the represented visual structures. In other words, a transformation will be able to be decoded from such representations that well encode the visual structures before and after the transformation \cite{zhang2019aet}.



For this purpose, we present a novel paradigm of Autoencoding Variational Transformations (AVT) to learn powerful representations that equivary against a generic class of transformations. We formalize it from an information-theoretic perspective by considering a joint
probability between images and transformations. This enables us to
use the mutual information to characterize the dependence between the representations and the transformations. Then, the AVT model can be trained directly by maximizing the mutual information in an unsupervised fashion without any labels. This will ensure the resultant representations contain intrinsic information about the visual structures that could be transformed extrinsically for individual images. Moreover, we will show that the representations learned in this way can be computed directly from the transformations and the representations of the original images {\em without a direct access to the original samples}, and this allows us to generalize the existing linear Transformation Equivariant Representations (TERs) to more general nonlinear cases \cite{qi2019learning}.



Unfortunately, it is intractable to maximize the mutual information directly as it is impossible to  exactly evaluate the posterior of a transformation from the associated representation. Thus, we instead seek to maximize a variational lower bound of the mutual information by introducing a transformation decoder to approximate the intractable posterior.
This results in an efficient autoencoding transformation (instead of data) architecture by jointly encoding a transformed image and decoding the associated transformation.


The resultant AVT model disruptively differs from the conventional auto-encoders \cite{hinton1994autoencoders,japkowicz2000nonlinear,vincent2008extracting} that seek to learn representations by reconstructing images.
Although the transformation could be decoded from the reconstructed original and transformed images, this is a quite strong assumption as such representations could contain {\em more than enough} information about both necessary and unnecessary visual details. The AVT model is based on a weaker assumption that the representations are trained to contain only the {\em necessary} information about visual structures to decode the transformation between the original and transformed images.
Intuitively, it is harder to reconstruct a high-dimensional image than decoding a transformation that has fewer degrees of freedom. In this sense, conventional auto-encoders tend to over-represent an image with every detail, no matter if they are necessary or not. Instead, the AVT could learn more generalizable representations by identifying the most essential visual structures to decode transformations, thereby yielding better performances for downstream tasks.


This remainder of this paper is organized as follows. In Section~\ref{sec:related}, we will review the related works on unsupervised methods. We will formalize the proposed AVT model by maximizing the mutual information between representations and transformations in Section~\ref{sec:formulation}. It is followed by the variational approach elaborated in Section~\ref{sec:var}. Experiment results will be demonstrated in Section~\ref{sec:exp} and we conclude the paper in Section~\ref{sec:conc}.

\section{Related Works}\label{sec:related}
In this section, we will review some related methods for training transformation-equivariant representations, along with the other unsupervised models.

\subsection{Transformation-Equivariant Representations}
The study of transformation-equivariance can be traced back to the idea of training capsule nets \cite{sabour2017dynamic,hinton2011transforming,hinton2018matrix}, where the capsules are designed to equivary to various transformations with vectorized rather than scalar representations. However, there was a lack of explicit training mechanism to ensure the resultant capsules be of transformation equivariance. 

To address this problem, many efforts have been made in literature  \cite{cohen2016group,cohen2016steerable,lenssen2018group} to extend the conventional translation-equivariant convolutions to cover more transformations.
For example, group equivariant convolutions (G-convolution) \cite{cohen2016group} have been developed to equivary to more types of transformations so that a richer family of geometric structures can be explored by the classification layers on top of the generated representations.
The idea of group equivariance has also been introduced to the capsule nets \cite{lenssen2018group} by ensuring the equivariance of output pose vectors to a group of transformations with a generic routing mechanism.

However, these group equivariant convolutions and capsules must be trained in a supervised fashion \cite{cohen2016group,lenssen2018group} with labeled data for specific tasks, instead of learning unsupervised transformation-equivariant representations generalizable to unseen tasks. Moreover, their representations are restricted to be a function of groups, which limits the ability of training future classifiers on top of more flexible representations.

Recently, Zhang et al.~\cite{zhang2019aet} present a novel Auto-Encoding Transformation (AET) model by learning a representation from which an input transformation can be reconstructed. This is closely related to our motivation of learning transformation equivariant representations, considering the transformation can be decoded from the learned representation of original and transformed images. On the contrary, in this paper, we approach it from an information-theoretic point of view in a more principled fashion.


Specifically, we will define a joint probability over the representations and transformations, and this will enable us to train unsupervised representations by directly maximizing the mutual information between the transformations and the representations. We wish the resultant representations can generalize to new tasks without access to their labels beforehand.

\subsection{Other Unsupervised Representations}
{\noindent \bf Auto-Encoders and GANs.}
Training auto-encoders in an unsupervised fashion has been studied in literature \cite{hinton1994autoencoders,japkowicz2000nonlinear,vincent2008extracting}. Most auto-encoders are trained by minimizing the reconstruction errors on input {\em data} from the encoded representations.
A large category of auto-encoder variants have been proposed.
Among them is the Variational Auto-Encoder (VAE) \cite{kingma2013auto} that maximizes the lower-bound of the data likelihood to train a pair of probabilistic encoder and decoder, while beta-VAE seeks to disentangle representations by introducing an adjustable hyperparameter on the capacity of latent channel to balance between the independence constraint and the reconstruction accuracy \cite{higgins2017beta}.  Denoising auto-encoder \cite{vincent2008extracting}
seeks to reconstruct noise-corrupted data to
learn robust representation, while
contrastive Auto-Encoder \cite{rifai2011contractive} encourages to learn representations invariant to small perturbations on data.
Along this line, Hinton et al.~\cite{hinton2011transforming} propose capsule nets by minimizing the discrepancy between the reconstructed and target data.

Meanwhile, Generative Adversarial Nets (GANs) have also been used to train unsupervised representations in literature.  Contrary to the auto-encoders, a GAN model generates data from the noises drawn from a simple distribution, with a discriminator trained adversarially to distinguish between real and fake data. The sampled noises can be viewed as the representation of generated data over a manifold, and one can train an encoder by inverting the generator to find the generating noise. This can be implemented by jointly training a pair of mutually inverse generator and encoder \cite{donahue2016adversarial,dumoulin2016adversarially}. There also exist better generalizable GANs in producing unseen data based on the Lipschitz assumption on the real data distribution \cite{qi2017loss,arjovsky2017wasserstein}, which can give rise to more powerful representations of data out of training examples  \cite{donahue2016adversarial,dumoulin2016adversarially,edraki2018generalized}. Compared with the Auto-Encoders, GANs do not rely on learning one-to-one reconstruction of data; instead, they aim to generate the entire distribution of data.

{\noindent \bf Self-Supervisory Signals.} There exist many other unsupervised learning methods using different types of self-supervised signals to train deep networks.
Mehdi and Favaro~\cite{noroozi2016unsupervised} propose to solve Jigsaw puzzles to train a convolutional neural network.
Doersch et al.~\cite{doersch2015unsupervised} train the network by predicting the relative positions between sampled patches from an image as self-supervised information. Instead, Noroozi et al.~\cite{noroozi2017representation} count features that satisfy equivalence relations between downsampled and tiled images, while Gidaris et al.~\cite{gidaris2018unsupervised} classify a discrete set of image rotations to train deep networks. Dosovitskiy et al.~\cite{dosovitskiy2014discriminative} create a set of surrogate classes by applying various transformations to individual images. However, the resultant features could over-discriminate visually similar images as they always belong to different surrogate classes.
Unsupervised features have also been learned from videos by estimating the self-motion of moving objects between consecutive frames \cite{agrawal2015learning}.


\section{Formulation}\label{sec:formulation}
We begin with the notations for the proposed unsupervised learning of the transformation equivariant representations (TERs). Consider a random sample $\mathbf x$ drawn from the data distribution $p(\mathbf x)$. We sample a transformation $\mathbf t$ from a distribution $p(\mathbf t)$, and apply it to $\mathbf x$, yielding a transformed image $\mathbf t(\mathbf x)$.

Usually, we consider a distribution $p(\mathbf t)$ of parameterized transformations, e.g., affine transformations with the rotations, translations and shearing constants being sampled from a simple distribution, and projective transformations that randomly shift and interpolate four corners of images. Our goal is to learn an unsupervised representation that contains as much information as possible to recover the transformation. We wish such a representation is able to compactly encode images such that it could equivary as the visual structures of images are transformed.

Specifically, we seek to learn an encoder that maps a transformed sample $\mathbf t(\mathbf x)$ to the mean $f_\theta$ and variance $\sigma_\theta$ of a desired representation. This results in the following probabilistic representation $\mathbf z$ of $\mathbf t(\mathbf x)$:
\begin{equation}\label{eq:rep}
\mathbf z = f_\theta(\mathbf t(\mathbf x))+\sigma_\theta(\mathbf t(\mathbf x)) \circ \epsilon
\end{equation}
where $\epsilon$ is sampled from a normal distribution $\mathcal N(\epsilon|\mathbf 0, \mathbf I)$, and $\circ$ denotes the element-wise product. In this case, the probabilistic representation $\mathbf z$ follows a normal distribution $p_\theta(\mathbf z|\mathbf t, \mathbf x)\triangleq \mathcal N\left(\mathbf z|f_\theta(\mathbf t(\mathbf x)),\sigma_\theta^2(\mathbf t(\mathbf x))\right)$ conditioned on the randomly sampled transformation $\mathbf t$ and input data $\mathbf x$. Meanwhile, the representation $\mathbf {\tilde z}$ of the original sample $\mathbf x$ can be computed as a special case when $\mathbf t$ is set to an identity transformation.


As discussed in Section~\ref{sec:intro}, we seek to learn a representation $\mathbf z$ equivariant to the sampled transformation $\mathbf t$ whose information can be recovered as much as possible from the representation $\mathbf z$.  Thus, the most natural choice to formalize this notion of transformation equivariance is the mutual information $I(\mathbf t,\mathbf z|\mathbf {\tilde z})$ between $\mathbf z$ and $\mathbf t$ from an information-theoretic perspective. The larger the mutual information, the more knowledge about $\mathbf t$ can be inferred from the representation $\mathbf z$.

Moreover, it can be shown that the mutual information $I(\mathbf t;\mathbf z|\mathbf {\tilde z})$ is the lower bound of the joint mutual information $I(\mathbf z;(\mathbf t,\mathbf {\tilde z}))$ that attains its maximum value when $I(\mathbf z;\mathbf x|\mathbf {\tilde z},\mathbf t)=0$. In this case, $\mathbf x$ provides no additional information about $\mathbf z$ once $(\mathbf {\tilde z},\mathbf t)$ are given. This implies one can estimate $\mathbf z$ directly from $(\mathbf {\tilde z},\mathbf t)$ without accessing the original sample $\mathbf x$, which generalizes the {\em linear} transformation equivariance to {\em nonlinear} case. For more details, we refer the readers to the long version of this paper \cite{qi2019learning}.

Therefore, we maximize the mutual information between the representation and the transformation to train the model
$$
\max_{\theta} I(\mathbf t;\mathbf z|\mathbf {\tilde z})
$$

Unfortunately, this maximization problem requires us to evaluate the posterior $p_\theta(\mathbf t|\mathbf z,\mathbf {\tilde z})$ of the transformation, which is often difficult to compute directly. This makes it intractable to train the representation by directly maximizing the above mutual information. Thus, we will turn to a variational approach by introducing a transformation decoder $q_\phi(\mathbf t|\mathbf z,\mathbf {\tilde z})$ with the parameter $\phi$ to approximate $p_\theta(\mathbf t|\mathbf z,\mathbf {\tilde z})$. In the next section, we will elaborate on this variational approach.

\section{Autoencoding Variational Transformations}\label{sec:var}
First, we present a variational lower bound of the mutual information $I(\mathbf t;\mathbf z|\mathbf x)$ that can be maximized over $q_\phi$ in a tractable fashion.

Instead of lower-bounding data likelihood in other variational approaches such as variational auto-encoders \cite{kingma2013auto}, it is more natural for us to maximize the lower bound of the mutual information \cite{agakov2004algorithm} between the representation $\mathbf z$ and the transformation $\mathbf t$ in the following way
\[
\begin{aligned}
&I(\mathbf t;\mathbf z|\mathbf {\tilde z}) = H(\mathbf t|\mathbf {\tilde z}) - H(\mathbf t|\mathbf z,\mathbf {\tilde z})\\
&=H(\mathbf t|\mathbf {\tilde z}) +\mathop\mathbb E\limits_{p_\theta(\mathbf t,\mathbf z,\mathbf {\tilde z})} \log p_\theta(\mathbf t|\mathbf z,\mathbf {\tilde z})\\
&= H(\mathbf t|\mathbf {\tilde z})  + \mathop\mathbb E\limits_{p_\theta(\mathbf t,\mathbf z,\mathbf {\tilde z})} \log q_\phi(\mathbf t|\mathbf z,\mathbf {\tilde z})\\
&+ \mathop\mathbb E\limits_{p(\mathbf z,\mathbf {\tilde z})} D(p_\theta(\mathbf t|\mathbf z,\mathbf {\tilde z})\|q_\phi(\mathbf t|\mathbf z,\mathbf {\tilde z}))\\
&\geq H(\mathbf t|\mathbf {\tilde z})  + \mathop\mathbb E\limits_{p_\theta(\mathbf t,\mathbf z,\mathbf {\tilde z})} \log q_\phi(\mathbf t|\mathbf z,\mathbf {\tilde z}) \triangleq \tilde I_{\theta,\phi}(\mathbf t;\mathbf z|\mathbf {\tilde z})
\end{aligned}
\]
where $H(\cdot)$ denotes the (conditional) entropy, and $D(p_\theta(\mathbf t|\mathbf z,\mathbf {\tilde z})\|q_\phi(\mathbf t|\mathbf z,\mathbf {\tilde z}))$ is the Kullback divergence between $p_\theta$ and $q_\phi$, which is always nonnegative.

We choose to maximize the lower variational bound $\tilde I(\mathbf t;\mathbf z|\mathbf {\tilde z})$. Since $H(\mathbf t|\mathbf {\tilde z})$ is independent of the model parameters $\theta$ and $\phi$, we simply maximize
\begin{equation}\label{eq:var}
\max\limits_{\theta,\phi}  \mathop\mathbb E\limits_{p_\theta(\mathbf t,\mathbf z,\mathbf{\tilde z})}\log q_\phi(\mathbf t|\mathbf z,\mathbf {\tilde z})
\end{equation}
to learn $\theta$ and $\phi$ under the expectation over $p(\mathbf t, \mathbf z, \mathbf {\tilde z})$.

This variational approach differs from the variational auto-encoders \cite{kingma2013auto}: the latter attempts to lower bound the data loglikelihood, while we instead seek to lower bound the mutual information here. Although both are derived based on an auto-encoder structure, the mutual information has a simpler form of lower bound than the data likelihood -- it does not contain an additional Kullback-Leibler divergence term, and thus shall be easier to maximize.

\subsection{Algorithm}
In practice, given a batch of samples $\{\mathbf x^{i}|i=1,\cdots,n\}$, we first draw a  transformation $\mathbf t^i$ corresponding to each sample.
Then we use the reparameterization (\ref{eq:rep}) to generate the probabilistic representation $\mathbf z^i$ with $f_\theta$ and $\sigma_\theta$ as well as a sampled noise $\epsilon^i$.

On the other hand, we use a normal distribution $\mathcal N(\mathbf t|d_\phi(\mathbf z,\mathbf {\tilde z}),\sigma^2_\phi(\mathbf z,\mathbf {\tilde z}))$ as the decoder $q_\phi(\mathbf t|\mathbf z,\mathbf {\tilde z})$, where the mean $d_\phi(\mathbf z,\mathbf {\tilde z})$ and variance $\sigma^2_\phi(\mathbf z, \mathbf {\tilde z})$ are implemented by deep network respectively.

With the above samples, the objective (\ref{eq:var}) can be approximated as
\begin{equation}\label{eq:obj}
\max\limits_{\theta,\phi} \dfrac{1}{n} \sum_{i=1}^n \log \mathcal N(\mathbf t^i|d_\phi(\mathbf z^i,\mathbf {\tilde z}^i),\sigma_\phi(\mathbf z^i,\mathbf {\tilde z}^i))
\end{equation}
where
\[
\mathbf z^i = f_\theta(\mathbf t^i(\mathbf x^i))+\sigma_\theta(\mathbf t^i(\mathbf x^i)) \circ \epsilon^i
\]
and
\[
\mathbf {\tilde z}^i = f_\theta(\mathbf x^i)+\sigma_\theta(\mathbf x^i) \circ \tilde\epsilon.
\]
and $\epsilon^i, \tilde\epsilon^i \sim \mathcal N(\epsilon|\mathbf 0, \mathbf I)$, and $\mathbf t^i\sim p(\mathbf t)$ for each $i=1,\cdots,n$.

\subsection{Architecture}

\begin{figure}[t]
    \centering
    \begin{subfigure}[c]{0.49\textwidth}
        \includegraphics[width=\textwidth]{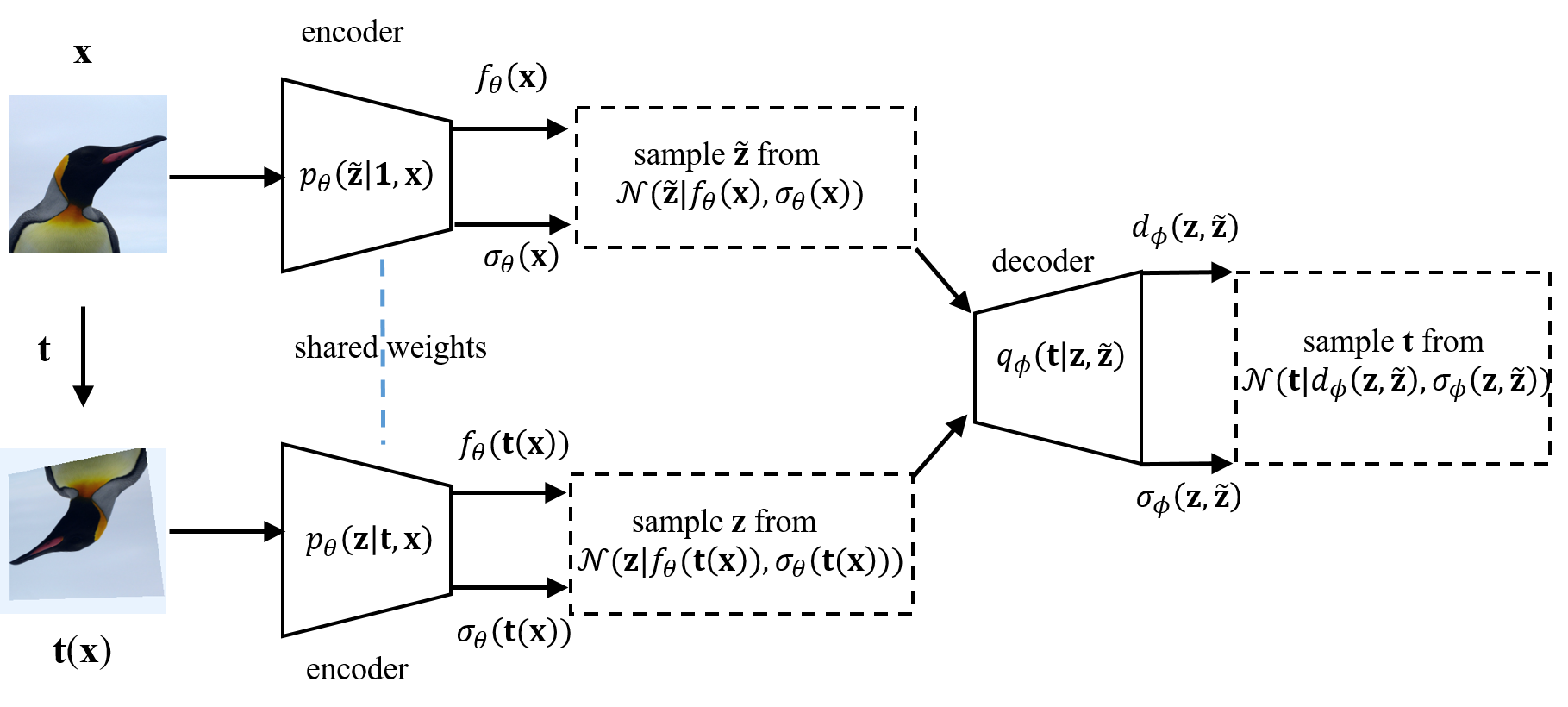}
    \end{subfigure}\\\vspace{-4mm}
    \caption{The architecture of the proposed AVT. The original and transformed images are fed through the encoder $p_\theta$ where $\mathbf 1$ denotes an identity transformation to generate the representation of the original image. The resultant representations $\mathbf {\tilde z}$ and $\mathbf z$ of original and transformed images are sampled and fed into the transformation decoder $q_\phi$ from which the transformation $\mathbf t$ is sampled. }\label{fig:avt}
\end{figure}

As illustrated in Figure~\ref{fig:avt}, we implement the transformation decoder $q_\phi(\mathbf t|\mathbf z,\mathbf {\tilde z})$ by using
a Siamese encoder network with shared weights to represent the original and transformed images with $\mathbf{\tilde z}$ and $\mathbf z$ respectively, where the mean $d_\phi$ and the variance $\sigma_\phi^2$ of the sampled transformation are predicted from the concatenation of both representations.




We note that, in a conventional auto-encoder, error signals must be backpropagated through a deeper decoder to reconstruct images before they train the encoder of interest. In contrast, the AVT allows a shallower decoder to estimate transformations with fewer variables so that stronger training signals can reach the encoder before it attenuates remarkably. This can more sufficiently train the encoder to represent images in downstream tasks.

%

\vspace{-1mm}
\section{Experiments}\label{sec:exp}
\vspace{-1mm}
In this section, we evaluate the proposed AVT model by following the standard protocol in literature.

\vspace{-1mm}
\subsection{CIFAR-10 Experiments}
\vspace{-1mm}
We evaluate the AVT model on the CIFAR-10 dataset.
\vspace{-2mm}
\subsubsection{Experiment Settings}
\vspace{-1mm}


\vspace{2mm}
{\noindent \bf Architecture} For a fair comparison with existing models, the Network-In-Network (NIN) is adopted on the CIFAR-10 dataset for the unsupervised learning task \cite{zhang2019aet}.  
The NIN consists of four convolutional blocks, each containing three convolutional layers.
The AVT has two NIN branches, each of which takes the original and transformed images as its input, respectively. We average-pool and concatenate the output feature maps from the forth block of two branches to form a $384$-d feature vector. Then an output layer follows to output the mean $d_\phi$ and the log-of-variance $\log \sigma_\phi^2$ of the predicted transformation, with the logarithm scaling the variance to a real value.

The two branches share the same network weights, with the first two blocks of each branch being used as the encoder network to directly output the mean $f_\theta$ of the representation. An additional $1\times 1$ convolution followed by a batch normalization layer is added on top of the representation mean to output the log-of-variance $\log\sigma_\theta^2$.

\vspace{2mm}
{\noindent\bf Implementation Details} The AVT networks are trained by the SGD with a batch size of $512$ images and their transformed versions. Momentum and weight decay are set to $0.9$ and $5\times 10^{-4}$, respectively. The model is trained for a total of $4,500$ epochs. The learning rate is initialized to $10^{-3}$. Then it is gradually decayed to $10^{-5}$ from $3,000$ epochs after it is increased to $5\times 10^{-3}$ after the first $50$ epochs.
The previous research \cite{zhang2019aet} has shown the projective transformation outperforms the affine transformation in training unsupervised models, and thus we adopt it to train the AVT for a fair comparison.
The projective transformation is composed of a
random translation of the four corners of an image in both horizontal and vertical directions by $\pm 0.125$ of its height and width, after it is scaled by a factor of $[0.8, 1.2]$ and rotated randomly with an angle from $\{0^\circ, 90^\circ, 180^\circ, 270^\circ\}$.

During training the AVT model, a single representation is randomly sampled from the encoder $p_\theta(\mathbf z|\mathbf t,\mathbf x)$, which is fed into the surrogate decoder $q_\phi(\mathbf t|\mathbf x, \mathbf z)$. In contrast, to fully exploit the uncertainty of probabilistic representations in training the downstream classification tasks, five random samples are drawn and averaged as the representation of an image used by the classifiers. We found averaging randomly sampled representations outperforms only using the mean of the representation to train the downstream classifiers.


\vspace{-1mm}
\subsubsection{Results}
\vspace{-1mm}

{\noindent\bf Comparison with Other Methods.} A classifier is usually trained upon the representation learned by an unsupervised model to assess the performance. Specifically, on CIFAR-10, the existing evaluation protocol \cite{oyallon2015deep,dosovitskiy2014discriminative,radford2015unsupervised,oyallon2017scaling,gidaris2018unsupervised,zhang2019aet} is strictly followed by building a classifier on top of the second convolutional block.

First, we evaluate the classification results by using the AVT features with both model-based and model-free classifiers.  For the model-based classifier, we follow \cite{zhang2019aet} by training a non-linear classifier with three Fully-Connected (FC) layers -- each of the two hidden layers has $200$ neurons with batch-normalization and ReLU activations, and the output layer is a soft-max layer with ten neurons each for an image class. We also test a convolutional classifier upon the unsupervised features by adding a third NIN block whose output feature map is averaged pooled and connected to a linear soft-max classifier.

\begin{table}
\caption{Comparison between unsupervised feature learning methods on CIFAR-10. The fully supervised NIN and the random Init. + conv have the same three-block NIN architecture, but the first is fully supervised while the second is trained on top of the first two blocks that are randomly initialized and stay frozen during training.}\label{tab01}
\centering
 \begin{tabular}{l|c} \toprule
Method&Error rate\\ \midrule
Supervised NIN \cite{gidaris2018unsupervised} (Upper Bound)&7.20  \\
Random Init. + conv \cite{gidaris2018unsupervised} (Lower Bound)&27.50  \\ \midrule
Roto-Scat + SVM \cite{oyallon2015deep} &17.7 \\
ExamplarCNN \cite{dosovitskiy2014discriminative} &15.7 \\
DCGAN \cite{radford2015unsupervised}&17.2 \\
Scattering \cite{oyallon2017scaling}&15.3\\
RotNet + non-linear \cite{gidaris2018unsupervised}&10.94\\
RotNet + conv \cite{gidaris2018unsupervised}&8.84\\
AET-affine + non-linear \cite{zhang2019aet} &9.77\\
AET-affine + conv \cite{zhang2019aet} &8.05\\
AET + non-linear \cite{zhang2019aet} &9.41\\
AET + conv \cite{zhang2019aet} &7.82\\ \midrule
AVT + non-linear &\textbf{8.96} \\
AVT + conv &\textbf{7.75} \\
\bottomrule
\end{tabular}
\end{table}

Table~\ref{tab01} shows the results by the AVT and other models. It compares the AVT with both fully supervised and unsupervised methods on CIFAR-10. The unsupervised AVT with the convolutional classifier almost achieves the same error rate as its fully supervised NIN counterpart with four convolutional blocks ($7.75\%$ vs. $7.2\%$). This remarkable result demonstrates the AVT could greatly close the performance gap with the fully supervised model on CIFAR-10.

\begin{table}
\caption{Error rates of different classifiers trained on top of the learned representations on CIFAR 10, where $n$-FC denotes a classifier with $n$ fully connected layers and conv denotes the third NIN block as a convolutional classifier. Two AET variants are chosen for a fair direct comparison since they are based on the same architecture as the AVT and have outperformed the other unsupervised representations before \cite{zhang2019aet}.}\label{tab02}
\centering
 \begin{tabular}{c|cccc} \toprule
   &1 FC&2 FC&3 FC&conv\\ \midrule
AET-affine \cite{zhang2019aet} &17.16 &9.77 &10.16 &8.05\\
AET-project \cite{zhang2019aet}&16.65 &9.41 &9.92 &7.82 \\ \midrule
(Ours) AVT &\textbf{16.19} &\textbf{8.96} &\textbf{9.55} &\textbf{7.75}\\\bottomrule
\end{tabular}
\end{table}

We also evaluate the AVT when varying numbers of FC layers and a convolutional classifier are trained on top of unsupervised representations respectively in Table~\ref{tab02}.
The results show that AVT can consistently achieve the smallest errors no matter which classifiers are used.



\begin{table}
\caption{The comparison of the KNN error rates by different models with varying numbers $K$ of nearest neighbors on CIFAR-10.}\label{tab04}
\centering
\small
 \begin{tabular}{c|ccccc} \toprule
$K$   &3&5&10&15&20\\ \midrule
AET-affine \cite{zhang2019aet}&24.88&23.29&23.07&23.34&23.94\\ \
AET-project \cite{zhang2019aet}&23.29&22.40&\bf 22.39&23.32&23.73 \\
(Ours) AVT &\bf 22.46&\bf 21.62&23.7&\bf 22.16&\bf 21.51 \\ \bottomrule
\end{tabular}
\end{table}

{\noindent\bf Comparison based on Model-free KNN Classifiers.}  We also test the model-free KNN classifier based on the averaged-pooled feature representations from the second convolutional block. The KNN classifier is model-free without training a classifier from labeled examples.  This enables us to make a direct evaluation on the quality of learned features.
Table~\ref{tab04} reports the KNN results with varying numbers of nearest neighbors.  Again, the AVT outperforms the compared representations when they are used to calculate $K$ nearest neighbors for classifying images.

\begin{table*}
\caption{Error rates on CIFAR-10 when different numbers of samples per class are used to train the downstream classifiers. A third convolutional block is trained with the labeled examples on top of the first two blocks of the NIN ($*$ the 13-layer network) pre-trained with the unlabeled data.
We compare with the fully supervised models that are trained with all the labeled examples from scratch.}\label{tab03}
\centering
 \begin{tabular}{c|ccccc} \toprule
   &20&100&400&1000&5000\\ \midrule
Supervised conv &66.34&52.74 &25.81 &16.53 &6.93\\
Supervised non-linear &65.03&51.13 &27.17 &16.13 &7.92\\\midrule
RotNet + conv \cite{gidaris2018unsupervised}&35.37 &24.72&17.16&13.57 &8.05 \\
AET-project + conv \cite{zhang2019aet} &\bf 34.83&24.35 &16.28 &12.58 &7.82 \\
AET-project + non-linear \cite{zhang2019aet} &\bf 37.13&25.19&18.32 & 14.27 &9.41 \\
\midrule
AVT + conv &35.44&\textbf{24.26} &\textbf{15.97} &\textbf{12.27} &\textbf{7.75}\\
AVT + non-linear &37.62&\textbf{25.01} &\textbf{17.95} &\textbf{14.14} &\textbf{8.96}\\\midrule
AVT + conv (13 layers)$^*$ &\bf 26.2&\bf 18.44&\bf 13.56&\bf 10.86 &\bf 6.3\\\bottomrule
\end{tabular}
\end{table*}

{\noindent\bf Comparison with Small Labeled Data.} Finally, we also conduct experiments when a small number of labeled examples are used to train the downstream classifiers on top of the learned representations.
This will give us some insight into how the unsupervised representations could help with only few labeled examples.   Table~\ref{tab03} reports the results of different models on CIFAR-10. The AVT outperforms the fully supervised models when only a small number of labeled examples ($\leq 1000$ samples per class) are available. It also performs better than the other unsupervised models in most of cases. Moreover, if we adopt the widely used 13-layer network \cite{laine2016temporal} on CIFAR-10 to train the unsupervised and supervised parts, the error rates can be further reduced significantly particularly when very few labeled examples are used.


\vspace{-1mm}
\subsection{ImageNet Experiments}
\vspace{-1mm}
We further evaluate the performance by AVT on the ImageNet dataset. The AlexNet is used as the backbone to learn the unsupervised features.
\vspace{-1mm}
\subsubsection{Architectures and Training Details}
\vspace{-1mm}
Two AlexNet branches with shared parameters are created with original and transformed images as inputs respectively to train unsupervised AVT. The $4,096$-d output features from the second last fully connected layer in two branches are concatenated and fed into the output layer producing the mean and the log-of-variance of eight projective transformation parameters. We still use SGD to train the network, with a batch size of $768$ images and the transformed counterparts, a momentum of $0.9$, a weight decay of $5\times 10^{-4}$. The initial learning rate is set to $10^{-3}$, and it is dropped by a factor of $10$ at epoch 300 and 350. The AVT is trained for $400$ epochs in total. Finally, the projective transformations are randomly sampled in the same fashion as on CIFAE-10, and the unsupervised representations fed into the classifiers are the average over five sampled representations from the probabilistic encoder.
\vspace{-1mm}
\subsubsection{Results}
\vspace{-1mm}
\begin{table}
\caption{Top-1 accuracy with non-linear layers on ImageNet. AlexNet is used as backbone to train the unsupervised models. After unsupervised features are learned, nonlinear classifiers are trained on top of Conv4 and Conv5 layers with labeled examples to compare their performances. We also compare with the fully supervised models and random models that give upper and lower bounded performances. For a fair comparison, only a single crop is applied in AVT and no dropout or local response normalization is applied during the testing. }\label{tab05}
\centering
 \begin{tabular}{l|cc} \toprule
Method&Conv4 &Conv5\\ \midrule
Supervised from \cite{bojanowski2017unsupervised}(Upper Bound)&59.7&59.7  \\
Random from \cite{noroozi2016unsupervised} (Lower Bound)&27.1 &12.0  \\ \midrule
Tracking \cite{wang2015unsupervised} &38.8&29.8 \\
Context \cite{doersch2015unsupervised} &45.6&30.4 \\
Colorization \cite{zhang2016colorful}&40.7&35.2 \\
Jigsaw Puzzles \cite{noroozi2016unsupervised}&45.3&34.6\\
BIGAN \cite{donahue2016adversarial}&41.9&32.2\\
NAT \cite{bojanowski2017unsupervised}&-&36.0\\
DeepCluster \cite{caron2018deep} &-&44.0\\
RotNet \cite{gidaris2018unsupervised}&50.0&43.8\\
AET-project \cite{zhang2019aet} &{53.2}&{47.0}\\\midrule
(Ours) AVT &\textbf{54.2}&\textbf{48.4}\\\bottomrule
\end{tabular}
\end{table}

\begin{table*}
\caption{Top-1 accuracy with linear layers on ImageNet. AlexNet is used as backbone to train the unsupervised models under comparison. A $1,000$-way linear classifier is trained upon various convolutional layers of feature maps that are spatially resized to have about $9,000$ elements. Fully supervised and random models are also reported to show the upper and the lower bounds of unsupervised model performances. Only a single crop is used and no dropout or local response normalization is used during testing for the AVT, except the models denoted with * where ten crops are applied to compare results.}\label{tab06}
\centering
 \begin{tabular}{l|ccccc} \toprule
Method&Conv1 &Conv2&Conv3&Conv4&Conv5\\ \midrule
ImageNet labels(Upper Bound)&19.3&36.3&44.2&48.3&50.5  \\
Random (Lower Bound)&11.6 &17.1&16.9&16.3&14.1  \\
Random rescaled \cite{krahenbuhl2015data}&17.5 &23.0&24.5&23.2&20.6  \\
\midrule
Context \cite{doersch2015unsupervised} &16.2&23.3&30.2&31.7&29.6 \\
Context Encoders \cite{pathak2016context}&14.1&20.7&21.0&19.8&15.5 \\
Colorization\cite{zhang2016colorful}&12.5&24.5&30.4&31.5&30.3\\
Jigsaw Puzzles \cite{noroozi2016unsupervised}&18.2&28.8&34.0&33.9&27.1\\
BIGAN \cite{donahue2016adversarial}&17.7&24.5&31.0&29.9&28.0\\
Split-Brain \cite{zhang2017split}&17.7&29.3&35.4&35.2&32.8\\
Counting \cite{noroozi2017representation}&18.0&30.6&34.3&32.5&25.7\\
RotNet \cite{gidaris2018unsupervised}&18.8&31.7&38.7&38.2&36.5\\
AET-project \cite{zhang2019aet}&19.2&32.8&40.6&39.7&37.7\\
\midrule
(Ours) AVT &\bf 19.5&\bf 33.6&\bf 41.3&\bf 40.3&\bf 39.1\\
\bottomrule
\toprule
DeepCluster* \cite{caron2018deep} &13.4&32.3&41.0&39.6&38.2\\
AET-project* \cite{zhang2019aet} &19.3&35.4&44.0&43.6&42.4\\
(Ours) AVT*&\textbf{20.9}&\textbf{36.1}&\textbf{44.4}&\textbf{44.3}&\textbf{43.5}\\\bottomrule
\end{tabular}
\end{table*}

\begin{table*}
\caption{Top-1 accuracy on the Places dataset. A $205$-way logistic regression classifier is trained on top of various layers of feature maps that are spatially resized to have about $9,000$ elements. All unsupervised features are pre-trained on the ImageNet dataset, and then frozen when training the logistic regression classifiers with Places labels. We also compare with fully-supervised networks trained with Places Labels and ImageNet labels, as well as with random models. The highest accuracy values are in bold and the second highest accuracy values are underlined.}\label{tab07}
\centering
 \begin{tabular}{l|ccccc} \toprule
Method&Conv1 &Conv2&Conv3&Conv4&Conv5\\ \midrule
Places labels(Upper Bound)\cite{zhou2014learning}&22.1&35.1&40.2&43.3&44.6 \\
ImageNet labels&22.7&34.8&38.4&39.4&38.7\\
Random (Lower Bound)&15.7 &20.3&19.8&19.1&17.5  \\
Random rescaled \cite{krahenbuhl2015data}&21.4 &26.2&27.1&26.1&24.0  \\
\midrule
Context \cite{doersch2015unsupervised} &19.7&26.7&31.9&32.7&30.9 \\
Context Encoders \cite{pathak2016context}&18.2&23.2&23.4&21.9&18.4 \\
Colorization\cite{zhang2016colorful}&16.0&25.7&29.6&30.3&29.7\\
Jigsaw Puzzles \cite{noroozi2016unsupervised}&\underline{23.0}&31.9&35.0&34.2&29.3\\
BIGAN \cite{donahue2016adversarial}&22.0&28.7&31.8&31.3&29.7\\
Split-Brain \cite{zhang2017split}&21.3&30.7&34.0&34.1&32.5\\
Counting \cite{noroozi2017representation}&\textbf{23.3}&\textbf{33.9}&36.3&34.7&29.6\\
RotNet \cite{gidaris2018unsupervised}&21.5&31.0&35.1&34.6&33.7\\
 AET-project \cite{zhang2019aet}& 22.1&32.9&\underline{37.1}&\underline{36.2}&\underline{34.7}\\\midrule
AVT&22.3&\underline{33.1}&\textbf{37.8}&\textbf{36.7}&\textbf{35.6}\\\bottomrule
\end{tabular}
\end{table*}

Table~\ref{tab05} reports the Top-1 accuracies of the compared methods on ImageNet by following the evaluation protocol in \cite{noroozi2016unsupervised,zhang2017split,gidaris2018unsupervised,zhang2019aet}. Two settings are adopted for evaluation, where Conv4 and Conv5 mean to train the remaining part of AlexNet on top of Conv4 and Conv5 with the labeled data. All the bottom convolutional layers up to Conv4 and Conv5 are frozen after they are trained in an unsupervised fashion. 
From the results, in both settings, the AVT model consistently outperforms the other unsupervised models.

We also compare with the fully supervised models that give the upper bound of the classification performance by training the whole AlexNet with all labeled data end-to-end. The classifiers of random models are trained on top of Conv4 and Conv5 whose weights are randomly sampled, which set the lower bounded performance. By comparison, the AVT model further closes the performance gap to the full supervised models to $5.5\%$ and $11.3\%$ on Conv4 and Conv5 respectively. This is a relative improvement by $15\%$ and $11\%$ over the previous state-of-the-art AET model.


Moreover, we also follow the testing protocol adopted in \cite{zhang2019aet} to compare the models by training a $1,000$-way linear classifier on top of different numbers of convolutional layers in Table~\ref{tab06}.  Again, the AVT consistently outperforms all the compared unsupervised models in terms of the Top-1 accuracy.
\vspace{-1mm}
\subsection{Places Experiments}
\vspace{-1mm}
Finally, we evaluate the AVT model on the Places dataset. Table~\ref{tab07} reports the results. Unsupervised models are pretrained on the ImageNet dataset, and a linear logistic regression classifier is trained on top of different layers of convolutional feature maps with Places labels. It assesses the generalizability of unsupervised features from one dataset to another. The models are still based on AlexNet variants. We compare with the fully supervised models trained with the Places labels and ImageNet labels respectively, as well as with the random networks. The AVT model outperforms the other unsupervised models, except performing slightly worse than Counting \cite{zhang2017split} with a shallow representation by Conv1 and Conv2.
\vspace{-1mm}
\section{Conclusion}\label{sec:conc}
\vspace{-1mm}
In this paper, we present a novel paradigm of learning representations by Autoencoding Variational Transformations (AVT) instead of reconstructing data as in conventional autoencoders. It aims to maximize the mutual information between the transformations and the representations of transformed images. The intractable maximization problem on mutual information is solved by introducing a transformation decoder to approximate the posterior of transformations through a variational lower bound. This naturally leads to a new probabilistic structure with a representation encoder and a transformation decoder. The resultant representations should contain as much information as possible about the transformations to equivary with them. Experiment results show the AVT representations set new state-of-the-art performances on CIFAR-10, ImageNet and Places datasets, greatly closing the performance gap to the supervised models as compared with the other unsupervised models.


{\small
\bibliographystyle{ieee}
\bibliography{egbib,aet_egbib}

\begin{thebibliography}{10}\itemsep=-1pt

\bibitem{agakov2004algorithm}
D.~B.~F. Agakov.
\newblock The im algorithm: a variational approach to information maximization.
\newblock {\em Advances in Neural Information Processing Systems}, 16:201,
  2004.

\bibitem{agrawal2015learning}
P.~Agrawal, J.~Carreira, and J.~Malik.
\newblock Learning to see by moving.
\newblock In {\em Proceedings of the IEEE International Conference on Computer
  Vision}, pages 37--45, 2015.

\bibitem{arjovsky2017wasserstein}
M.~Arjovsky, S.~Chintala, and L.~Bottou.
\newblock Wasserstein gan.
\newblock {\em arXiv preprint arXiv:1701.07875}, 2017.

\bibitem{bojanowski2017unsupervised}
P.~Bojanowski and A.~Joulin.
\newblock Unsupervised learning by predicting noise.
\newblock {\em arXiv preprint arXiv:1704.05310}, 2017.

\bibitem{caron2018deep}
M.~Caron, P.~Bojanowski, A.~Joulin, and M.~Douze.
\newblock Deep clustering for unsupervised learning of visual features.
\newblock {\em arXiv preprint arXiv:1807.05520}, 2018.

\bibitem{cohen2016group}
T.~Cohen and M.~Welling.
\newblock Group equivariant convolutional networks.
\newblock In {\em International conference on machine learning}, pages
  2990--2999, 2016.

\bibitem{cohen2018intertwiners}
T.~S. Cohen, M.~Geiger, and M.~Weiler.
\newblock Intertwiners between induced representations (with applications to
  the theory of equivariant neural networks).
\newblock {\em arXiv preprint arXiv:1803.10743}, 2018.

\bibitem{cohen2016steerable}
T.~S. Cohen and M.~Welling.
\newblock Steerable cnns.
\newblock {\em arXiv preprint arXiv:1612.08498}, 2016.

\bibitem{doersch2015unsupervised}
C.~Doersch, A.~Gupta, and A.~A. Efros.
\newblock Unsupervised visual representation learning by context prediction.
\newblock In {\em Proceedings of the IEEE International Conference on Computer
  Vision}, pages 1422--1430, 2015.

\bibitem{donahue2016adversarial}
J.~Donahue, P.~Kr{\"a}henb{\"u}hl, and T.~Darrell.
\newblock Adversarial feature learning.
\newblock {\em arXiv preprint arXiv:1605.09782}, 2016.

\bibitem{dosovitskiy2014discriminative}
A.~Dosovitskiy, J.~T. Springenberg, M.~Riedmiller, and T.~Brox.
\newblock Discriminative unsupervised feature learning with convolutional
  neural networks.
\newblock In {\em Advances in Neural Information Processing Systems}, pages
  766--774, 2014.

\bibitem{dumoulin2016adversarially}
V.~Dumoulin, I.~Belghazi, B.~Poole, O.~Mastropietro, A.~Lamb, M.~Arjovsky, and
  A.~Courville.
\newblock Adversarially learned inference.
\newblock {\em arXiv preprint arXiv:1606.00704}, 2016.

\bibitem{edraki2018generalized}
M.~Edraki and G.-J. Qi.
\newblock Generalized loss-sensitive adversarial learning with manifold
  margins.
\newblock In {\em Proceedings of European Conference on Computer Vision (ECCV
  2018)}, 2018.

\bibitem{gidaris2018unsupervised}
S.~Gidaris, P.~Singh, and N.~Komodakis.
\newblock Unsupervised representation learning by predicting image rotations.
\newblock {\em arXiv preprint arXiv:1803.07728}, 2018.

\bibitem{higgins2017beta}
I.~Higgins, L.~Matthey, A.~Pal, C.~Burgess, X.~Glorot, M.~Botvinick,
  S.~Mohamed, and A.~Lerchner.
\newblock beta-vae: Learning basic visual concepts with a constrained
  variational framework.
\newblock In {\em International Conference on Learning Representations}, 2017.

\bibitem{hinton2011transforming}
G.~E. Hinton, A.~Krizhevsky, and S.~D. Wang.
\newblock Transforming auto-encoders.
\newblock In {\em International Conference on Artificial Neural Networks},
  pages 44--51. Springer, 2011.

\bibitem{hinton2018matrix}
G.~E. Hinton, S.~Sabour, and N.~Frosst.
\newblock Matrix capsules with em routing.
\newblock 2018.

\bibitem{hinton1994autoencoders}
G.~E. Hinton and R.~S. Zemel.
\newblock Autoencoders, minimum description length and helmholtz free energy.
\newblock In {\em Advances in neural information processing systems}, pages
  3--10, 1994.

\bibitem{japkowicz2000nonlinear}
N.~Japkowicz, S.~J. Hanson, and M.~A. Gluck.
\newblock Nonlinear autoassociation is not equivalent to pca.
\newblock {\em Neural computation}, 12(3):531--545, 2000.

\bibitem{kingma2013auto}
D.~P. Kingma and M.~Welling.
\newblock Auto-encoding variational bayes.
\newblock {\em arXiv preprint arXiv:1312.6114}, 2013.

\bibitem{krahenbuhl2015data}
P.~Kr{\"a}henb{\"u}hl, C.~Doersch, J.~Donahue, and T.~Darrell.
\newblock Data-dependent initializations of convolutional neural networks.
\newblock {\em arXiv preprint arXiv:1511.06856}, 2015.

\bibitem{krizhevsky2012imagenet}
A.~Krizhevsky, I.~Sutskever, and G.~E. Hinton.
\newblock Imagenet classification with deep convolutional neural networks.
\newblock In {\em Advances in neural information processing systems}, pages
  1097--1105, 2012.

\bibitem{laine2016temporal}
S.~Laine and T.~Aila.
\newblock Temporal ensembling for semi-supervised learning.
\newblock {\em arXiv preprint arXiv:1610.02242}, 2016.

\bibitem{lenssen2018group}
J.~E. Lenssen, M.~Fey, and P.~Libuschewski.
\newblock Group equivariant capsule networks.
\newblock {\em arXiv preprint arXiv:1806.05086}, 2018.

\bibitem{noroozi2016unsupervised}
M.~Noroozi and P.~Favaro.
\newblock Unsupervised learning of visual representations by solving jigsaw
  puzzles.
\newblock In {\em European Conference on Computer Vision}, pages 69--84.
  Springer, 2016.

\bibitem{noroozi2017representation}
M.~Noroozi, H.~Pirsiavash, and P.~Favaro.
\newblock Representation learning by learning to count.
\newblock In {\em The IEEE International Conference on Computer Vision (ICCV)},
  2017.

\bibitem{oyallon2017scaling}
E.~Oyallon, E.~Belilovsky, and S.~Zagoruyko.
\newblock Scaling the scattering transform: Deep hybrid networks.
\newblock In {\em International Conference on Computer Vision (ICCV)}, 2017.

\bibitem{oyallon2015deep}
E.~Oyallon and S.~Mallat.
\newblock Deep roto-translation scattering for object classification.
\newblock In {\em Proceedings of the IEEE Conference on Computer Vision and
  Pattern Recognition}, pages 2865--2873, 2015.

\bibitem{pathak2016context}
D.~Pathak, P.~Krahenbuhl, J.~Donahue, T.~Darrell, and A.~A. Efros.
\newblock Context encoders: Feature learning by inpainting.
\newblock In {\em Proceedings of the IEEE Conference on Computer Vision and
  Pattern Recognition}, pages 2536--2544, 2016.

\bibitem{qi2017loss}
G.-J. Qi.
\newblock Loss-sensitive generative adversarial networks on lipschitz
  densities.
\newblock {\em arXiv preprint arXiv:1701.06264}, 2017.

\bibitem{qi2019learning}
G.-J. Qi.
\newblock Learning generalized transformation equivariant representations via
  autoencoding transformations.
\newblock {\em arXiv preprint arXiv:1906.08628}, 2019.

\bibitem{radford2015unsupervised}
A.~Radford, L.~Metz, and S.~Chintala.
\newblock Unsupervised representation learning with deep convolutional
  generative adversarial networks.
\newblock {\em arXiv preprint arXiv:1511.06434}, 2015.

\bibitem{rifai2011contractive}
S.~Rifai, P.~Vincent, X.~Muller, X.~Glorot, and Y.~Bengio.
\newblock Contractive auto-encoders: Explicit invariance during feature
  extraction.
\newblock In {\em Proceedings of the 28th International Conference on
  International Conference on Machine Learning}, pages 833--840. Omnipress,
  2011.

\bibitem{sabour2017dynamic}
S.~Sabour, N.~Frosst, and G.~E. Hinton.
\newblock Dynamic routing between capsules.
\newblock In {\em Advances in Neural Information Processing Systems}, pages
  3856--3866, 2017.

\bibitem{vincent2008extracting}
P.~Vincent, H.~Larochelle, Y.~Bengio, and P.-A. Manzagol.
\newblock Extracting and composing robust features with denoising autoencoders.
\newblock In {\em Proceedings of the 25th international conference on Machine
  learning}, pages 1096--1103. ACM, 2008.

\bibitem{wang2015unsupervised}
X.~Wang and A.~Gupta.
\newblock Unsupervised learning of visual representations using videos.
\newblock In {\em Proceedings of the IEEE International Conference on Computer
  Vision}, pages 2794--2802, 2015.

\bibitem{zhang2019aet}
L.~Zhang, G.-J. Qi, L.~Wang, and J.~Luo.
\newblock Aet vs. aed: Unsupervised representation learning by auto-encoding
  transformations rather than data.
\newblock {\em arXiv preprint arXiv:1901.04596}, 2019.

\bibitem{zhang2017split}
R.~Zhang, P.~Isola, and A.~A. Efros.
\newblock Split-brain autoencoders: Unsupervised learning by cross-channel
  prediction.

\bibitem{zhang2016colorful}
R.~Zhang, P.~Isola, and A.~A. Efros.
\newblock Colorful image colorization.
\newblock In {\em European Conference on Computer Vision}, pages 649--666.
  Springer, 2016.

\bibitem{zhou2014learning}
B.~Zhou, A.~Lapedriza, J.~Xiao, A.~Torralba, and A.~Oliva.
\newblock Learning deep features for scene recognition using places database.
\newblock In {\em Advances in neural information processing systems}, pages
  487--495, 2014.

\end{thebibliography}
}

\end{document}